\documentclass[10pt,twocolumn,letterpaper]{article}

\usepackage{iccv}              

\definecolor{iccvblue}{rgb}{0.21,0.49,0.74}
\usepackage[pagebackref,breaklinks,colorlinks,allcolors=iccvblue]{hyperref}

\def\paperID{*****} 
\def\confName{ICCV}
\def\confYear{2025}

\title{QR-LoRA: QR-Based Low-Rank Adaptation for Efficient Fine-Tuning of Large Language Models}

\author{Jessica E. Liang\\
University of Pennsylvania\\
{\tt\small jeliang@seas.upenn.edu}
\and
Anirudh Bharadwaj\\
University of Pennsylvania\\
{\tt\small anirudh2@seas.upenn.edu}
}

\begin{document}
\maketitle

\begin{abstract}
The growing scale of Large Language Models (LLMs) has necessitated the development of parameter-efficient fine-tuning techniques. Low-Rank Adaptation (LoRA) has emerged as a promising approach, reducing the number of trainable parameters by applying low-rank updates to pretrained weights. While standard LoRA learns both update factors directly, several recent variants first initialize those matrices via an SVD of the pretrained weights—an operation that can be expensive on large models and yields singular vectors that are not always easy to interpret. In this work, we extract an orthonormal basis from the pretrained weight matrix using QR decomposition with column pivoting, and then express the LoRA update as a linear combination of these basis vectors—training only the scalar coefficients, which imposes clear structure on adaptation and drastically reduces parameter count. Experiments across GLUE tasks show that QR-LoRA matches or exceeds the performance of full fine-tuning, standard LoRA, and SVD-LoRA (LoRA with update matrices initialized via singular value decomposition) with as few as 601 parameters—a reduction of over $1000\times$ compared to full fine-tuning and $77\times$ fewer than typical LoRA setups.
\end{abstract}

\section{Introduction}

The rapid proliferation of Large Language Models (LLMs) has revolutionized NLP, enabling breakthroughs in tasks from machine translation to question answering. However, fine-tuning these models end-to-end requires updating hundreds of millions, or even billions, of parameters, incurring substantial computational, storage, and environmental costs. Resource‑constrained settings such as on‑device personalization motivate methods that adapt large models with minimal parameter overhead while preserving accuracy.

Existing parameter-efficient adapters such as Low-Rank Adaptation (LoRA) decompose each weight update \(\Delta W\) into a product \(BA\), reducing trainable parameters from \(O(d^2)\) to \(O(rd)\) \citep{hu2022lora}. SVD-based variants further compress by selecting singular vectors, but require expensive decompositions per matrix and produce bases without an inherent notion of importance ordering. Consequently, even state-of-the-art adapters often demand tens of thousands of parameters to match full fine-tuning performance.

In this work, we propose QR-LoRA, a novel adapter that constructs an orthonormal basis via pivoted QR decomposition of each frozen weight matrix. Pivoting naturally orders basis vectors by the magnitudes of the diagonal entries of \(R\), yielding an interpretable ranking of directions. We then parameterize the update as
\[
\Delta W \;=\; \sum_{i=1}^{r} \lambda_{i}\,Q_{i}R_{i}^{T},
\]
training only the scalar coefficients \(\{\lambda_{i}\}\). This approach fixes the basis \(Q\) and upper-triangular factors \(R\), requiring as few as 601 trainable parameters for a RoBERTa-base model while retaining rich representational capacity.

We evaluate multiple QR-LoRA configurations on eight tasks from the GLUE benchmark~\citep{wang2019gluemultitaskbenchmarkanalysis}, comparing against full fine-tuning, standard LoRA, and SVD-LoRA. The smallest variant (which trains only 601 parameters) matches or outperforms full fine-tuning on four tasks and standard LoRA on five tasks, while training $1000\times$ fewer parameters than full fine-tuning and $77\times$ fewer than LoRA. These results demonstrate that pivoted QR bases enable highly parameter-efficient adaptation without sacrificing accuracy.

In Appendix \ref{app:1}, we provide some related work on LoRA and matrix decomposition.

\section{Background}
\label{app:2}

In this section, we review the two main ingredients underlying QR-LoRA: low-rank adapters (LoRA) and the QR decomposition with thresholding.

\subsection{Low-Rank Adapters (LoRA)}
LoRA~\cite{hu2022lora} injects trainable low-rank update matrices into transformer weight layers while keeping the majority of the pre-trained parameters frozen.  Concretely, for a weight matrix \(W\in\mathbb{R}^{d\times d}\), LoRA parameterizes the update as
\[
  \Delta W = BA,
  \quad
  B\in\mathbb{R}^{d\times r},\;
  A\in\mathbb{R}^{r\times d},\;
  r\ll d,
\]
where only \(A\) and \(B\) are learned.  This reduces the number of trainable parameters from \(\mathcal{O}(d^2)\) to \(\mathcal{O}(r\,d)\), often recovering most of the full fine-tuning performance at a small fraction of the cost.  Subsequent work has explored alternative factorization schemes (e.g.\ SVD-LoRA, QLoRA) and adaptive rank selection (AdaLoRA).

\subsection{QR Decomposition with Thresholding}
The QR decomposition factorizes a matrix \(W\in\mathbb{R}^{d\times d}\) into an orthonormal factor \(Q\) and an upper‐triangular factor \(R\):
\begin{equation}
      W = QR.
\end{equation}
When combined with column‐pivoting or magnitude‐based thresholding, QR yields a truncated basis that retains the most “energetic” directions of \(W\).  Given a threshold \(\tau\in(0,1)\), one can select the smallest \(k\) such that
\(\sum_{i=1}^{k}\lvert R_{ii}\rvert \ge \tau\,\sum_{j=1}^{d}\lvert R_{jj}\rvert\),
and use only the first \(k\) columns of \(Q\) (and corresponding rows of \(R\)) for downstream updates.  This produces an adaptive low-rank approximation that is both orthonormal and interpretable, and can be computed efficiently even on large weight matrices.

\section{Methodology: QR-LoRA}

\subsection{Low-Rank Update via Orthonormal Basis}
We propose modifying LoRA by introducing a QR-based adaptation mechanism. 
Given a pretrained weight matrix $W_0 \in \mathbb{R}^{L \times M}$, we first compute its reduced QR decomposition~\cite{strang2022introduction}:
\begin{equation}
    W_0 = Q R,
\end{equation}
where $Q \in \mathbb{R}^{L \times L}$ is an orthonormal matrix and $R \in \mathbb{R}^{L \times M}$ is upper triangular. 
We propose to use
 QR decomposition with column pivoting~\cite{golub2013matrix}, so it reorders the columns during the decomposition process so that the magnitudes of the diagonal entries of 
$R$ are arranged in non-increasing order, thereby aligning the ``importance'' of components with their order in the decomposition. That is, the diagonal elements satisfy \( R_{11} \geq R_{22} \geq \cdots \geq R_{MM} \).

We define the low-rank update as:
\begin{equation}
    \Delta W = \sum_{i=1}^{r} \lambda_i Q_i R_i^T,
\end{equation}
where $Q_i$ is the $i$-th column of $Q$, $R_i^{T}$ is the \emph{column vector} obtained by transposing the $i$‑th row of $R$, and $\lambda_i$ are trainable scalars. This construction ensures that $\Delta W$ maintains the same dimensions as $W_0$, while dramatically reducing the number of trainable parameters.

We can
compute the cumulative energy (e.g., the sum of squared diagonal entries) and choose the smallest $r$ such that~\cite{candes2011robust} \begin{equation}
\frac{\sum_{i=1}^{r} R_{ii}^2}{\sum_{i=1}^{M} R_{ii}^2} \geq \tau,
\end{equation}
where $\tau$ is a threshold (say, 90–95\%).
This way, $r$ is selected based on how much of the ``information'' in $W_0$ is captured. For example, in RoBERTa-Base, $M=768$. When we used $\tau=0.5$ and apply QR-LoRA for $W_q$ in the last transformer layer, we have $r=150$.

\subsection{Theoretical Motivation and Impact of Orthonormality}

The use of an orthonormal basis $Q$ as the foundation for adaptation provides several important theoretical and practical benefits. First, orthonormal columns in $Q$ guarantee that each learned direction is independent and non-redundant, improving numerical conditioning and ensuring stable gradients~\cite{trefethen97}. 

Moreover, by limiting $\Delta W$ to a low-dimensional, fixed orthonormal subspace acts as a strong regularizer, potentially reducing overfitting. This connects to recent work on the intrinsic dimension of fine-tuning~\cite{aghajanyan2020intrinsicdimensionalityexplainseffectiveness, li2018measuring}, where restricting parameter updates to a small subspace was shown to improve generalization, especially in data-rich regimes. Additionally, the magnitude of each $R_{ii}$ provides a clear interpretation of the importance of each basis direction, facilitating principled rank selection and offering insights into the underlying structure of $W_0$.

While SVD produces singular vectors ordered by singular values (optimal for matrix approximation in the least-squares sense), QR with column pivoting provides a computationally efficient, interpretable alternative for orthonormal basis construction. QR is particularly attractive for very large matrices where full SVD is prohibitive, and has seen wide use in numerical linear algebra and signal processing~\cite{golub2013matrix, strang2022introduction}.

Taken together, these properties endow QR-LoRA with both strong theoretical grounding and significant practical advantages for efficient, robust adaptation of large-scale neural networks.

\section{Experiments}

\begin{table*}[ht]
\centering
\begin{tabular}{llccc}
\toprule
Category & Configuration & \# of Trainable P & Accuracy-1 (\%) & Accuracy-2 (\%) \\
\midrule
Fine‑tuning            & 3 + 5 epochs    &     125M                &  81.99      &  82.17        \\
Original LoRA          & $\Delta W = BA$, $r=2$  &     92,160        &  81.96     &  82.22       \\
SVD‑LoRA               & $r=2, k=1, \alpha=2$       &         46,080                & 80.14  & 80.48  \\
\midrule
QR‑LoRA       & $\tau=0.5$, all 12 layers $W_o$ &  1,702        &   82.05     &  \textbf{82.29}       \\
                       & $\tau=0.7$, all 12 layers $W_o$  &   3,142      &   82.04     &    82.25     \\
                       & $\tau=0.8$, all 12 layers $W_o$ &    4,053       &   \textbf{82.07}     &   82.28      \\
\midrule
QR‑LoRA  & $\tau=0.5$, last 4 layers $W_o$ &   614       &   81.99     &   82.19      \\
                       & $\tau=0.5$, last 4 layers $W_q$, $W_v$ &  1,311        &  81.98      &     82.22    \\
                       & $\tau=0.5$, all 12 layers $W_o$ &   1,702       &   82.05     &   82.29      \\
\bottomrule
\end{tabular}
\caption{Overview of experimental runs on MNLI}
\label{tab:MNLI}
\end{table*}


\begin{table*}[t]
\centering
\begin{tabular}{llccc}
\toprule
Category & Configuration & \# of Trainable P & Accuracy (\%) & F1 (\%) \\
\midrule
Fine‑tuning            & 3 + 5 epochs      &    125M               &   87.99    &    91.42     \\
Original LoRA          & $\Delta W = BA$, $r=2$  &    92,160         &   88.97    &  87.00       \\
SVD‑LoRA               & $r=2, k=1, \alpha=2$      &         46,080                 & 87.75  &  91.20  \\
\midrule
QR‑LoRA ($\tau$ sweep)      & $\tau=0.5$, all 12 layers $W_o$ &  1,702        &  88.73      &   91.96      \\
                       & $\tau=0.7$, all 12 layers $W_o$   &  3,142      &     88.73   &      91.96   \\
                       & $\tau=0.8$, all 12 layers $W_o$ &  4,053        &   88.73     &    91.96    \\
\midrule
QR‑LoRA (layer sweep)  & $\tau=0.5$, last 4 layers, $W_o$ &    614       &    \textbf{88.97}    &   \textbf{92.15}      \\
                       & $\tau=0.5$, last 4 layers, $W_q$, $W_v$& 1,311  &   88.73    &  91.96       \\
                       & $\tau=0.5$, all 12 layers, $W_o$ & 1,702   &     88.73   &   91.96      \\
\bottomrule
\end{tabular}
\caption{Overview of experimental runs on MRPC}
\label{tab:MRPC}
\end{table*}


\subsection{Experiment Setup}
We evaluate our approach on a subset of the GLUE benchmark~\citep{wang2019gluemultitaskbenchmarkanalysis}, specifically using the tasks MNLI, MRPC, SST-2, CoLA, QNLI, QQP, RTE, and STS-B. For each task, we train on up to $\min(10000,|\text{train}|)$ examples, ensuring consistency in data scale across experiments.

All methods use RoBERTa-base (125M parameters) as the starting point, which is first warm-up fine-tuned for three epochs. For the baseline comparisons, we include full fine-tuning (FT), in which all parameters are updated, and standard LoRA, where we freeze the transformer and learn a low-rank update $\Delta W = BA$ with $B \in \mathbb{R}^{d \times r}$, $A \in \mathbb{R}^{r \times d}$, and $r=2$. We also implement SVD-LoRA, which maintains the same rank ($r=2$), but initializes $B$ and $A$ using the top-$k$ singular vectors of each weight matrix, with scaling by $\alpha/r$.


\paragraph{QR‑LoRA configurations.}
QR‑LoRA treats the backbone encoder as frozen and injects a
low‑rank adapter into selected attention projections.
For each weight matrix \(W\in\{W_q,W_k,W_v,W_o\}\) chosen for
adaptation, we compute a pivoted‑QR decomposition
\(W = QR\).  Let \(r\) be the number of diagonal entries
of \(R\) whose magnitude exceeds \(\tau\,R_{11}\);
we keep the corresponding \(r\) columns of \(Q\) as an
orthonormal basis \(B\) and learn a diagonal coefficient matrix
\(A\in\mathbb{R}^{r\times r}\), updating \(W\leftarrow W + BA\).
All other parameters of the transformer remain frozen.

We explore three axes of variation:

\begin{itemize}
    \item [1.]
    \textbf{Threshold}: \(\tau\in\{0.5,0.7,0.8\}\), controlling the
   retained rank \(r\).
    \item [2.]
    \textbf{Adapter scope}: adapters inserted in (i) the final four
   attention blocks or (ii) all twelve blocks of RoBERTa‑base.
    \item [3.]
    \textbf{Projection set}: adapting  
   (a) output projections \(W_o\) only,  
   (b) the pair \((W_q,W_v)\), or  
   (c) all three \((W_q,W_v,W_o)\).
\end{itemize}

Layers and projection matrices not selected for adaptation are
left unchanged.

\begin{table*}[h]
\centering
  \begin{tabular}{@{}lccccccccc@{}}
    \toprule
    Method    & \# of Trainable P    & MNLI   & SST-2 & MRPC  & CoLA  & QNLI  & QQP   & RTE   & STS-B \\
    \midrule
    QR-LoRA1  & 1,311  & \textbf{82.10}  & \textbf{94.84} & 88.73 & 59.57 & 92.75 & 91.36 & 73.29 & 89.53 \\
    QR-LoRA2  &   601  & 82.09  & 94.72 & 88.73 & 59.82 & 92.77 & 91.36 & 72.56 & 89.47 \\
    SVD-LoRA  &46,080  & 80.31  & 91.97 & 87.75 & \textbf{61.58} & 87.73 & 85.07 & 67.51 & 90.15 \\
    LoRA      &92,160  & 82.09  & \textbf{94.84} & \textbf{89.71} & 58.59 & 92.66 & 91.40 & 72.20 & 89.87 \\
    FT        &125M    & 81.67  & 93.12 & 87.99 & 57.35 & \textbf{92.79} & \textbf{91.66} & \textbf{78.34} & \textbf{90.94} \\
    \bottomrule
  \end{tabular}
  \caption{Performance Comparison Across Different Methods: QR-LoRA1: (Wq, Wv, last 4, $\tau=0.5$), QR-LoRA2: (Wq only, last 4, $\tau=0.5$), SVD-LoRA: ($r=2, k=1, \alpha=2$), LoRA: ($\Delta W=BA$, $r=2$), and Fine-tuning (FT). 
}
\label{tab:all_datasets}
\end{table*}

\subsection{QR-LoRA Performance}

In Tables \ref{tab:MNLI} and \ref{tab:MRPC}, we summarize the performance of our experiments for MNLI and MRPC respectively.
In Table \ref{tab:MNLI}, \emph{Accuracy-1} is ``Matched Accuracy'' which
refers to evaluation on the same genre as the training data.
For example, if the training data includes fiction, government, and travel, the matched set contains test examples from these same genres.
\emph{Accuracy-2} is the ``Mismatched Accuracy''
which refers to evaluation on different genres than those seen during training.

\paragraph{Results on MNLI and MRPC.}
Across the configurations we explored, QR‑LoRA attains up to
82.07\,\% matched / 82.29\,\% mismatched accuracy on MNLI and
92.15\,\% F1 on MRPC while requiring at most
1311 trainable parameters (roughly \(10^{-3}\) \% of the model).
These numbers are within 0.1–0.3 pp of, and in several cases
slightly above, the 125M‑parameter full‑tuning baseline
(Table~\ref{tab:MNLI}, Table~\ref{tab:MRPC}).
In Appendix, Figure~\ref{fig:parameter_vs_performance} visualizes the resulting
parameter–performance trade‑off where QR‑LoRA has the lowest parameter count—among the
methods evaluated.
Similarly, on MRPC, QR‑LoRA not only matches the overall accuracy of full fine‑tuning (87.99\%) but also surpasses it by obtaining a top F1 score of 92.15\% with only 614 parameters—a clear improvement over the 91.42\% F1 achieved by fine‑tuning and the results from other lightweight variants. In Figure \ref{fig:parameter_vs_performance}, we clearly see how QR-LoRA outperforms full fine-tuning, original LoRA, and SVD-LoRA in both accuracy and F1 score. We can also notice that beyond 600 parameters, adding more (e.g. adapting $W_v$ or $W_o$) yields no additional gains.


We also evaluated two QR-LoRA configurations across all 8 GLUE tasks and compared them to standard fine-tuning and SVD-LoRA baselines (see results in table \ref{tab:all_datasets}). QR-LoRA1 tunes $W_q$ and $W_v$ in the last 4 attention layers (with $\tau=0.5$, and $1311$ total parameters), while QR-LoRA2 tunes only $W_q$ in the last 4 attention layers (with $\tau=0.5$, and $601$ total parameters).

QR-LoRA1 improves on standard fine-tuning on a variety of the additional GLUE tasks tested. In particular, QR-LoRA1 outperforms standard fine-tuning by 1.72 points on SST-2 (94.84 vs. 93.12), 1.72 points on MRPC (88.73 vs. 87.99), and 2.22 points on CoLA (59.57 vs. 57.35). Even on the subset of tasks where QR-LoRA1 does not do better than standard fine-tuning (QNLI, QQP, and SST-2) for the most part it is consistently competitive, coming within 1.5 points on each of these tasks. 

QR-LoRA2 is able to remain competitive with QR-LoRA1, coming within 0.7 points of QR-LoRA1's performance on all tasks. It is also able to outperform both SVD-LoRA ($\approx 46,000$ parameters) and standard LoRA ($\approx 92,000$ parameters) on some tasks, despite the large parameter discrepancy ($\approx 77\times$ fewer than SVD-LoRA, and $\approx 153\times$ fewer than LoRA).  

One notable discrepancy however, was in the results for the RTE task, for which standard fine-tuning did far better than all 4 of the other methods tested (QR-LoRA1, QR-LoRA2, SVD-LoRA, and LoRA), with over a 5 point gap to the second best technique, which was not observed for each of the other tasks, which saw far tighter clustering among the top techniques. We hypothesize that this is due to both the inherent difficulty of the RTE task (which necessitates quantifying entailment between two short sentences across a variety of domains) and the small number of examples afforded (at 2.5K training examples, this was, by some distance, the smallest of the 8 tasks we looked at). 

In Appendix \ref{app:3}, we provide more experimental results on training-set size ablation study.

\section{Conclusions and Future Work}
In this paper, we propose a novel QR-LoRA method with high efficiency in terms of the number of trainable parameters. We update only scalar coefficients $\lambda_i$ on a fixed orthonormal basis, allowing us to achieve an extremely low number of trainable parameters compared to Fine-Tuning (FT), LoRA, and SVD-LoRA (LoRA with update matrices initialized via singular value
decomposition) while also improving interpretability.  Parameter-efficient fine-tuning with QR decomposition shows strong scalability across data regimes. Many areas remain of interest for future work.

One limitation is that we only tested on the GLUE benchmark, which limits the impact of our results (in particular, assessing how well the observed performance gains from QR-LoRA generalize beyond these settings). Future work could be extended to evaluate on more challenging benchmarks, such as SuperGLUE or generation-oriented datasets, as well as assessing different architectures (ex. decoder-only models like GPT-3 or multimodal transformers).

Additionally, we only applied QR-LoRA to the attention projection matrices ($W_q,W_k,W_v$). In principle, the same QR-based adaptation could be extended to other layer types (feed-forward network weight matrices, embedding layers, and output heads), representing a potential future work. 

Across both MNLI and MRPC, we observed that changes in key hyperparameters, such as the threshold $\tau$, number of layers tuned, and the weights adapted ($W_o, W_q, W_v$) lead to only marginal differences in performance (see Table \ref{tab:MNLI} and Table \ref{tab:MRPC}). While this could suggest that QR-LoRA is robust to hyperparameter choices, it might also mean that our current evaluation setup lacks the resolution to capture meaningful differences. 

\section*{Acknowledgment}

The authors would like to thank Professor Mark Yatskar at the University of Pennsylvania for his invaluable guidance on this work.

{
    \small
    \bibliographystyle{ieeenat_fullname}
    \bibliography{ICCV_main}
}

\newpage
\appendix

\section{Related Work}
\label{app:1}

\paragraph{Adapter and Low-Rank Fine-Tuning.}
Parameter-efficient fine-tuning has rapidly advanced in response to the growing scale of large language models, with a central theme being the development of adapter-based and low-rank techniques. LoRA pioneered an efficient fine-tuning approach by freezing most of a pre-trained model’s parameters and learning compact, low-rank updates represented as a product of small matrices $\Delta W = BA$, significantly reducing computational overhead while maintaining high performance~\citep{hu2022lora}. This foundational idea has inspired numerous variants and extensions. For example, QLoRA introduced quantization into the LoRA framework by first compressing model weights into a 4-bit representation and then applying low-rank adaptations directly within this quantized parameter space, pushing the limits of memory efficiency for adaptation~\citep{dettmers2023qlora}. AdaLoRA further improved on the original formulation by introducing dynamic rank allocation, adjusting the rank of updates across layers during training in response to singular value magnitudes, which increased both parameter efficiency and flexibility~\citep{zhang2023adalora}. NLoRA leveraged Nyström method-inspired sketches to provide more efficient initializations, thus improving scalability and reducing the computational burden of adapter modules~\citep{guo2025nlora}.

Beyond these, recent work has improved both efficiency and expressivity. ALORA, LoRA-XS, and OLoRA introduce adaptive allocation, extreme sparsity, and orthonormal update bases, respectively, to better manage capacity and cost~\citep{liu2024alora, balazy2024lora, buyukakyuz2024olora}. DoRA separates weight magnitude from direction via decomposed updates, improving parameter efficiency and training stability~\citep{liu2024dora}. Tracking-LoRA introduces dynamic subspace tracking from numerical linear algebra, continually updating the low-rank space during training for faster convergence and better scalability on large models~\citep{lin2024trackinglora}. SparseAdapter enforces structured sparsity within adapters, reducing active parameters without degrading performance~\citep{he2022sparseadapter}.

\paragraph{Matrix Decomposition for Compression.}
Low-rank adapters leverage the finding that task-specific updates lie in low-dimensional subspaces~\citep{li2018measuring, aghajanyan2020intrinsicdimensionalityexplainseffectiveness}. Classical techniques like SVD and PCA exploit this by truncating to leading components, offering theoretical guarantees but incurring high costs for large models~\citep{candes2011robust, strang2022introduction}. QR-LoRA reduces this overhead using thresholded QR decomposition with column pivoting to extract an ordered orthonormal basis, where the diagonal of $R$ ranks direction importance~\citep{golub2013matrix, trefethen97}. Rather than tuning basis vectors, we fine-tune linear combinations of top-ranked directions via a global threshold, framing QR as a lightweight tool for basis extraction. Beyond matrix methods, tensor decompositions such as Tucker and CP offer complementary compression strategies for future subspace-adaptive designs~\citep{kossaifi2019tnet, phan2020stable, lebedev2015speeding}.

\section{More Experimental Results}
\label{app:3}

\subsection{Parameter-Performance Trade‑off}

\begin{figure*}[h]
  \centering
  \includegraphics[scale=0.55]{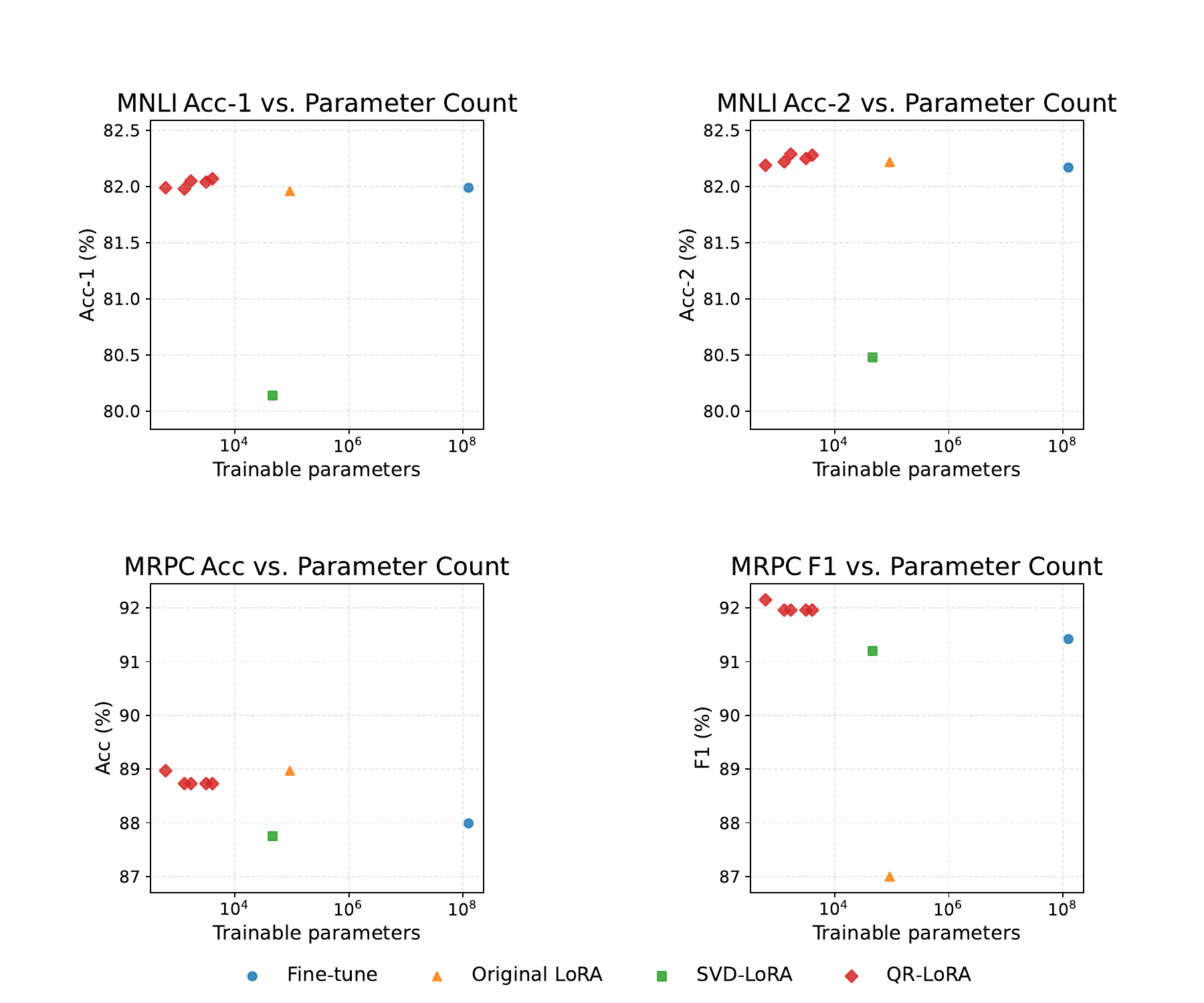}  
  \caption{Effect of trainable parameter count on downstream performance. Top row: MNLI matched (left) and mismatched (right) accuracy; bottom row: MRPC accuracy (left) and F1 (right), for Fine-tune, Original LoRA, SVD-LoRA and QR-LoRA variants.}
  \label{fig:parameter_vs_performance}
\end{figure*}

Figure~\ref{fig:parameter_vs_performance} visualizes the resulting
parameter–performance trade‑off: QR‑LoRA occupies the upper‑left
corner—highest accuracy at the lowest parameter count—among the
methods evaluated.

\subsection{Training‑Set Size Ablation}
\label{app:train-ablation}


\begin{table*}[h]
\centering
\smallskip
\begin{tabular}{@{}lllcc@{}}
\toprule
Configuration  & \# of Trainable P  & Training Data Size & Accuracy-1 (\%) & Accuracy-2 (\%) \\
\midrule
LoRA & 92,160  & 2{,}000   &   72.34   &  73.09    \\
QR-LoRA & 1,311       & 2{,}000  &  {72.39}    &  {73.50}    \\
FT &125M & 2{,}000  & \textbf{76.92} & \textbf{76.95}\\
\addlinespace
LoRA & 92,160  & 10{,}000 &   81.96   &   82.22   \\
QR-LoRA   & 1,311      & 10{,}000 &  {81.98}    &   \textbf{82.23}   \\
FT & 125M & 10{,}000 & \textbf{81.99}  & 82.17\\
\addlinespace
LoRA & 92,160  & 50{,}000 &    84.88  &   84.68   \\
QR-LoRA  & 1,311        & 50{,}000 &  \textbf{84.91}    &  \textbf{84.71}    \\
FT & 125M & 50{,}000 & 84.42 & 84.26\\
\bottomrule
\end{tabular}
\caption{Overall accuracy-1 (matched accuracy) and accuracy-2 (mismatched accuracy) for MNLI with LoRA variants under varying training data sizes. LoRA: ($\Delta W = BA$, $r=2$), QR-LoRA: (last 4 layers, $W_q, W_v$, $\tau=0.5$), FT: Fine Tuning.}
\label{tab:training_size_ablation}
\end{table*}

\paragraph{When does QR‑LoRA help?}
To better understand the performance characteristics of QR-LoRA, particularly in light of its underperformance on RTE (a low-resource, out-of-distribution task), we conducted a training set ablation study using the MNLI dataset. Our goal was to assess how QR-LoRA compares to standard LoRA and full fine-tuning (FT) under varying amounts of training data. Specifically, we evaluated all three methods on subsets of the MNLI training set with 2000, 10000, and 50000 examples. The results are shown in Table \ref{tab:training_size_ablation}.

We observe a pronounced shift in performance dynamics across the data regimes:

\textbf{2000 examples:} FT exceeds both LoRA variants by roughly four percentage points.  

\textbf{10000 examples:} QR‑LoRA and FT perform indistinguishably.  

\textbf{50000 examples:} QR‑LoRA achieves the best matched and mismatched accuracies.

These results indicate that QR‑LoRA is most advantageous in
moderate- to high‑resource settings where parameter efficiency is
desired, while FT remains preferable in the extreme low‑resource
regime. We hypothesize that this is due to the fact that, with little data, the few trainable coefficients in QR‑LoRA may under‑fit, whereas FT can still exploit its larger adjustment space; once the dataset passes $\approx10,000$ examples, the stronger implicit regularization of QR‑LoRA allows it to overtake FT. Extending this analysis to additional datasets and to tasks with substantial distribution shift (e.g.\ RTE) is left to future work.

\end{document}